\documentclass[letterpaper, 10pt, conference]{style/ieeeconf} 
\makeatletter
\let\NAT@parse\undefined
\makeatother

\usepackage{dblfloatfix}
\usepackage[numbers, sectionbib, sort]{natbib}
\usepackage{bm}
\usepackage{gensymb}
\usepackage{xcolor}
\usepackage{graphicx}
\usepackage{amsmath}
\usepackage{amssymb}
\usepackage{subcaption}
\usepackage{amsfonts}
\usepackage{siunitx}
\usepackage{booktabs}
\usepackage{makecell}
\usepackage{multirow}
\usepackage{upgreek}
\usepackage[font=small]{caption}
\usepackage[export]{adjustbox}
\usepackage{tikz}
\usepackage{tabularx}
\usepackage{sidecap} \sidecaptionvpos{figure}{c}
\usepackage[hidelinks]{hyperref}
\usepackage[nameinlink, capitalize]{cleveref}
\usepackage[printonlyused,withpage,nolist,nohyperlinks]{acronym}

\Crefname{section}{Sec.}{Sec.}
\Crefname{equation}{Eq.}{Eq.}

\IEEEoverridecommandlockouts

\title{\Large \bf Exploiting Priors from 3D Diffusion Models for RGB-Based One-Shot View Planning}

\author{Sicong Pan$^\star$ \and Liren Jin$^\star$ \and Xuying Huang \and Cyrill Stachniss \and Marija Popovi\'{c} \and Maren Bennewitz%
\thanks{$^\star$These authors contributed equally to this work.}
\thanks{All authors are with the University of Bonn, Germany. Cyrill Stachniss and Maren Bennewitz are additionally with the Lamarr Institute for Machine Learning and Artificial Intelligence, Bonn, Germany. This work has partially been funded by the Deutsche Forschungsgemeinschaft (DFG, German Research Foundation) under grant 459376902 – AID4Crops and under Germany’s Excellence Strategy, EXC-2070 – 390732324 – PhenoRob.
Corresponding: \texttt{span@uni-bonn.de}.}
}

\begin{document}

\maketitle
\thispagestyle{empty}
\pagestyle{empty}
\begin{abstract}
Object reconstruction is relevant for many autonomous robotic tasks that require interaction with the environment. A key challenge in such scenarios is planning view configurations to collect informative measurements for reconstructing an initially unknown object. One-shot view planning enables efficient data collection by predicting view configurations and planning the globally shortest path connecting all views at once. However, prior knowledge about the object is required to conduct one-shot view planning. In this work, we propose a novel one-shot view planning approach that utilizes the powerful 3D generation capabilities of diffusion models as priors. By incorporating such geometric priors into our pipeline, we achieve effective one-shot view planning starting with only a single RGB image of the object to be reconstructed. Our planning experiments in simulation and real-world setups indicate that our approach balances well between object reconstruction quality and movement cost.
\end{abstract}

\section{Introduction} \label{S:introduction}
Many autonomous robotic applications require 3D models of objects to perform downstream tasks, e.g., pose estimation~\citep{yang2022cvpr}, object manipulation~\citep{dengler2023iros}, and detection~\citep{zaenker2023iros}. When deployed in initially unknown environments, a robot often needs to reconstruct the objects before interacting with them. During this procedure, a challenge is planning a view sequence to acquire the most informative measurements to be integrated into the reconstruction system while minimizing the robot's travel distance or operation time.

Without any prior knowledge about the environment, a common strategy is to plan the next-best-view (NBV) iteratively based on the current reconstruction state~\citep{isler2016icra, sunderhauf2023icra, pan2022eccv, zeng2020iros, mendoza2020prl, menon2023iros, palazzolo2018drones}. However, NBV planning only generates a local path to the subsequent view and cannot effectively distribute the mission time or movement budget, resulting in suboptimal view planning performance. 
An alternative line of work considers one-shot view planning~\citep{pan2022ral1, pan2024icra, hu2024icra}. Given initial measurements of an object to be reconstructed, one-shot view planning predicts a set of views at once~and computes the globally shortest path connecting them. A robot's sensor then follows the planned path to collect measurements, which are used for object reconstruction after the data acquisition is completed. By decoupling data collection and object reconstruction, these approaches do not rely on iterative map updates for object-specific view planning online during a mission.
To perform one-shot view planning, prior knowledge about the object is required. Previous works consider planning priors based on multi-view images or partial point cloud observations. However, they either only handle a fixed view configuration~\citep{pan2024icra} or rely on depth sensors~\citep{hu2024icra, pan2022ral1}. 

\begin{figure}[!t]
\centering
 \begin{subfigure}[]{1.0\columnwidth}  \includegraphics[width=\columnwidth]{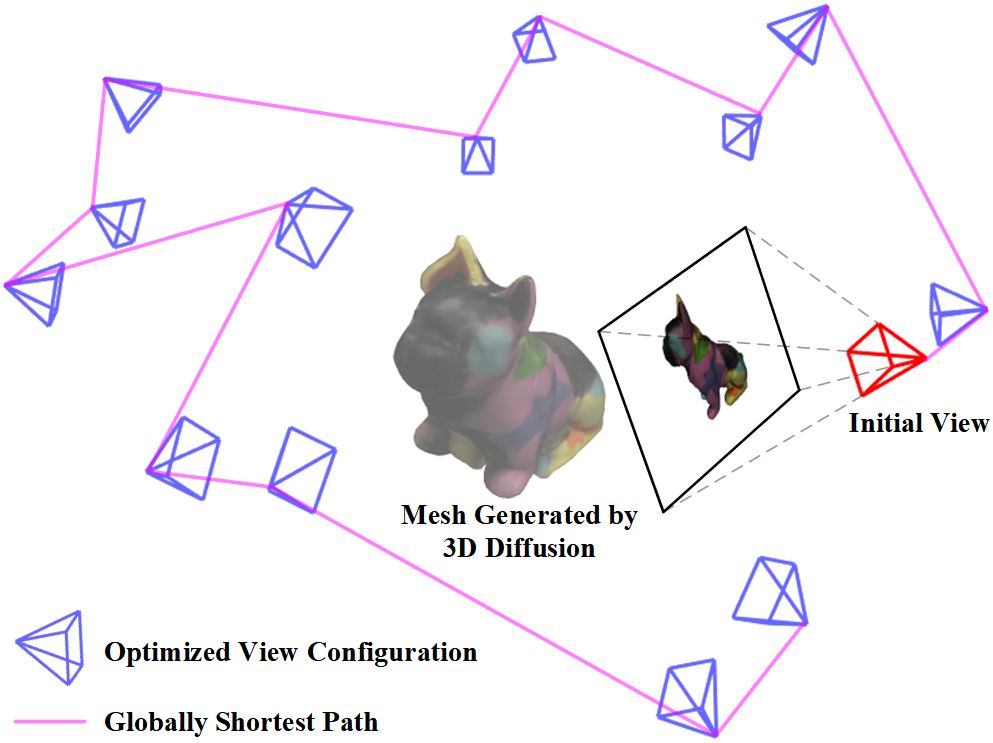}
 \end{subfigure}
   \caption{An example of our RGB-based one-shot view planning by exploiting priors from 3D diffusion models. Our goal is to plan a set of views (blue) at once to collect informative RGB images for object reconstruction. The key component in our approach is a 3D diffusion model generating the corresponding 3D mesh of a single RGB image from the initial camera view (red).
   By leveraging the mesh as geometric priors, our approach produces view configurations specifically associated with the target object and calculates the globally shortest path. In particular, we plan denser views to observe more geometrically complex parts (front part of the object in the example) to improve the reconstruction quality.}
   \label{F: teaser}
\vspace{-0.5cm}
\end{figure}

To address these aforementioned limitations, we propose integrating geometric priors from 3D diffusion models into one-shot view planning. Recently, 3D diffusion models emerged as a powerful tool for generating 3D content based on text prompts or a single image.
By training on large datasets, 3D diffusion models learn prior knowledge about objects commonly seen in real life~\citep{long2022eccv,long2023arxiv,liu2024neurips}. 
Humans similarly exploit prior knowledge to hallucinate 3D models of an object based on semantics and appearance information contained in RGB observations. However, recovering a 3D representation from a single RGB image is inherently an ill-posed problem and corresponds to multiple plausible solutions. 
As a result, models generated by 3D diffusion models do not reflect the exact representation of the object to be reconstructed. This prohibits their direct application as a method for accurate 3D representation, as required in robotics tasks. Incorporating the capabilities of 3D diffusion models to provide geometric priors in robotics remains an unexplored area.

The main contribution of this work is a novel RGB-based one-shot view planning approach that exploits the geometric priors from 3D diffusion models. Our approach enables view planning with an object-specific view configuration for object reconstruction as shown in Fig.~\ref{F: teaser}. 
A key component of our pipeline is a 3D diffusion model that outputs a 3D mesh of the object given one RGB image as input. This generated mesh is a proxy to the inaccessible ground truth 3D model and serves as the basis for our one-shot view planning. Given the generated 3D mesh, we convert the one-shot view planning into a customized set covering optimization problem to calculate the minimum set of views that densely covers the mesh, which we solve using linear programming. Our approach places the object-specific views and follows the globally shortest path for collecting informative RGB images around the object. After the data collection, we train a Neural Radiance Field (NeRF) using all collected images to acquire the object's 3D representation.

To the best of our knowledge, our approach is the first to leverage 3D diffusion models for view planning. We make the following claims: (i)~we exploit the powerful 3D diffusion models to enable our one-shot view planning starting with only one RGB image as input; (ii)~we design the one-shot view planning as a customized set covering optimization problem, yielding view configurations suitable for RGB-based object reconstruction using NeRFs. We conduct extensive experiments on publicly available object datasets and in real world scenarios, demonstrating the applicability and generalization ability of our approach. Our one-shot view planning allows for object-specific view placement to account for varying object geometries, achieving a better trade-off between movement cost and reconstruction quality compared to baselines. To support reproducibility and future research, our implementation is open-sourced at: \url{https://github.com/psc0628/DM-OSVP}

\section{Related Work} \label{S:related_work} 

In this section, we introduce relevant works on view planning for object reconstruction and diffusion models for 3D generation.

\subsection{View Planning for Object Reconstruction}
Object reconstruction is essential in many robotic applications. One important capability in this scenario is to actively reconstruct the object using a robot sensor. Without any prior knowledge, a common approach is to plan the NBV iteratively based on the current reconstruction state, thus maximizing the information of the object in a greedy manner. 
\citet{isler2016icra} propose selecting the NBV by calculating the information gain based on visibility and the likelihood of observing new parts of the object to be reconstructed. Similarly, Pan et al.~\cite{pan2022ral2,pan2023cviu} weight the 3D space based on visibility and distance to observe surfaces and then employ coverage optimization for NBV planning. In addition, \citet{menon2023iros} introduce a shape completion method based on partially observed objects and conduct NBV planning to cover the estimated missing surfaces. PC-NBV~\citep{zeng2020iros} trains a neural network to predict the utility of candidate views given partial point cloud observations. In the context of NBV planning for RGB-based object reconstruction, Jin et al.~\cite{jin2023iros} integrate uncertainty estimation into image-based neural rendering to guide NBV selection in a mapless way. Lin et al.~\citep{lin2022rssworkshop} and \citet{sunderhauf2023icra} train an ensemble of NeRF models, utilizing the ensemble's variance to measure uncertainty for NBV planning.

While showing promising object reconstruction results, NBV planning often relies on computationally intensive online map updates and its greedy nature leads to inefficient paths. To address these limitations, recent works propose one-shot view planning paradigm. Given an initial measurement, one-shot view planning predicts all required views at once and calculates the globally shortest path connecting them, resulting in reduced movement costs. 
The pioneering work SCVP~\citep{pan2022ral1} trains a neural network in a supervised way to directly predict the global view configuration given initial point cloud observations. To generate training labels, the authors solve the set covering problem to obtain a view configuration fully covering the ground truth 3D models. \citet{hu2024icra} further reduces the required views by incorporating a point cloud-based implicit surface reconstruction method to complete missing surfaces before conducting one-shot view planning. In the domain of RGB-based object reconstruction, \citet{pan2024icra} propose a view prediction network to predict the number of views to reconstruct an object using NeRFs required to reach its performance upper bound. However, due to the lack of geometric representations during the view planning stage, this work only considers distributing the views following a fixed pattern, without adapting view configurations to account for varying object geometries. 

Our work shares the same idea of using one-shot view planning to reconstruct an unknown object. Different from previous works that rely on depth sensors~\citep{hu2024icra, pan2022ral1} or fixed view configurations~\citep{pan2024icra}, our novel approach only requires RGB inputs and plans view configurations specifically associated with the objects, leading to better object reconstruction performance while reducing movement costs. 

\begin{figure*}[!t]
\centering
  \includegraphics[width=0.80\textwidth]{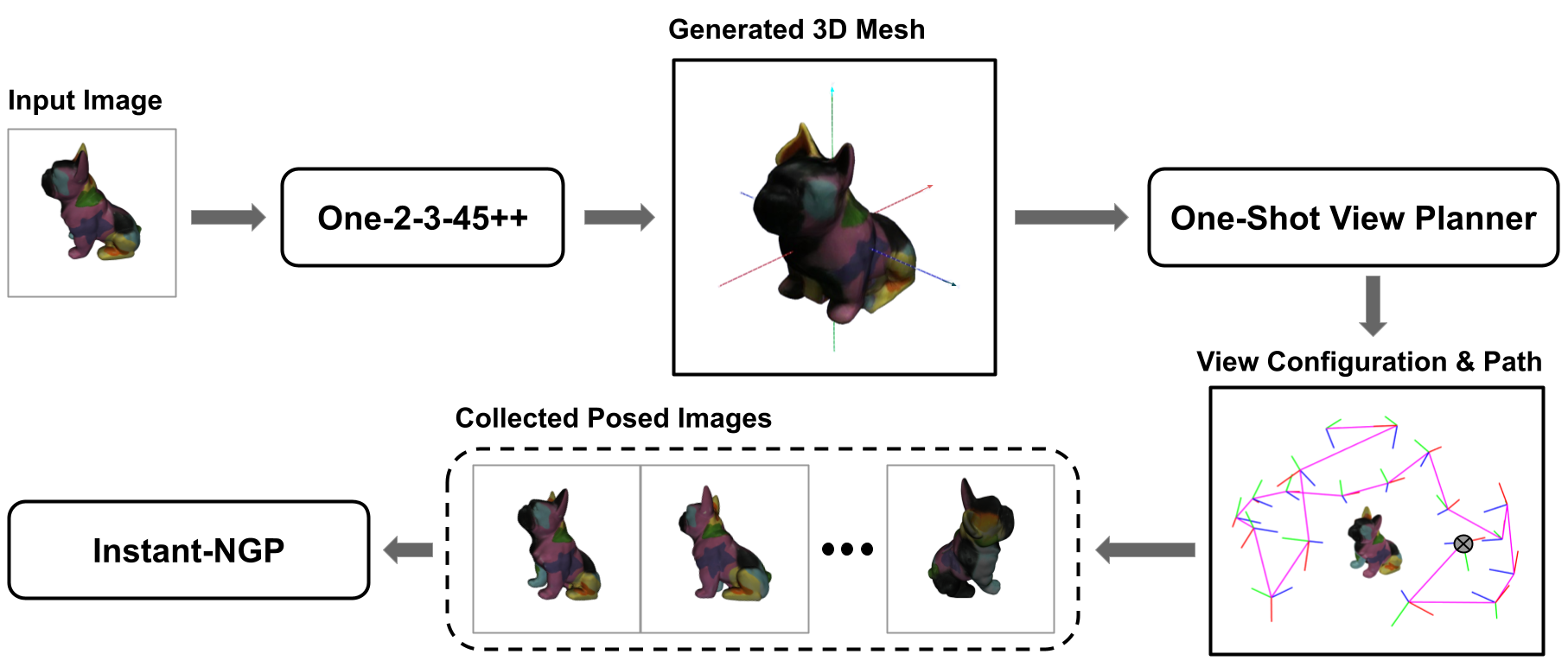}
  \caption{Overview of our proposed RGB-based one-shot view planning pipeline. Given a single RGB image of the object to be reconstructed, we leverage a 3D diffusion model, One-2-3-45++~\citep{liu2023arxiv}, to generate a 3D mesh. This mesh serves as a proxy to the ground truth geometry and is the basis for our view planning. Based on this prior, we construct the one-shot view planning task as a customized set covering optimization problem and solve it to obtain a minimum set of views required to densely cover the mesh surfaces. The RGB camera starts at the initial view (shown as $\otimes$) and follows the generated globally shortest path to collect RGB images, which we use to train a NeRF in Instant-NGP~\citep{muller2022tog} after the data acquisition is completed.}
  \label{F: framework}
\vspace{-0.5cm}
\end{figure*}

\subsection{Diffusion Models for 3D Generation}
Diffusion models are state-of-the-art generative models for producing plausible high-quality images. Starting from random Gaussian noises, diffusion models learn to subsequently denoise the input to finally recover the true images~\citep{ho2020neurips, rombach2022cvpr}. By training on large datasets, diffusion models acquire powerful prior knowledge and show their capabilities in the domain of 2D image generation.

Inspired by the advances of diffusion models, recent works investigate using diffusion models for 3D content generation. Given a text prompt describing a desired scene, DreamFusion~\citep{poole2023iclr}, ProlificDreamer~\citep{wang2023neurips}, and MVDream~\citep{shi2024iclr} optimize a differentiable 3D representation, e.g., NeRF, from scratch and leverage neural rendering to generate 2D images at different viewpoints. These rendered images are then fed into 2D diffusion models to calculate the similarity to the priors learned by the diffusion model, which guide the 3D shape optimization process. 
While showing impressive results, these methods suffer from prolonged rendering and optimization times, limiting their robotic applications.

Another line of work investigates fine-tuning pretrained 2D diffusion models for multi-view synthesis from single image inputs~\citep{liu2023iccv, liu2024iclr}. The follow-up work One-2-3-45~\citep{liu2024neurips} produces 3D meshes using images generated from the multi-view diffusion models. However, its performance is limited by the inconsistency between multi-view images. 
Recent 3D diffusion model One-2-3-45++~\citep{liu2023arxiv} mitigates the problem of inconsistencies by conditioning the multi-view image generation on each other. The generated multi-view consistent images are exploited as the guidance for 3D diffusion to directly produce high-quality meshes in a short time, i.e., within $60$\,s. In this work, we utilize geometric priors from 3D diffusion models to enable RGB-based one-shot view planning for object reconstruction.

\section{Our Approach} \label{S:our_approach}

We propose a novel RGB-based one-shot view planning method for unknown object reconstruction. An overview of our approach is shown in \cref{F: framework}. Given a single RGB measurement of the object, we leverage a 3D diffusion model to generate its corresponding mesh. Based on rich prior information contained in the generated mesh, we formulate one-shot view planning as a set covering optimization problem, which we solve with linear programming to acquire the minimum set of views densely covering mesh surfaces. We calculate the globally shortest path connecting all views for data collection using a robot's RGB camera. After data collection, we train a NeRF model using all collected RGB images to generate a 3D representation of the object.  

\subsection{Geometric Priors from 3D Diffusion Model}

A key component of our approach is a 3D diffusion model for predicting the corresponding mesh given only one RGB image as an initial observation. Specifically, we use the state-of-the-art 3D diffusion model One-2-3-45++~\citep{liu2023arxiv} for generating plausible meshes due to its accurate mesh generation and efficient inference compared to other 3D diffusion models~\citep{poole2023iclr, wang2023neurips, shi2024iclr}. One-2-3-45++ model is trained on Objaverse~\citep{deitke2023cvpr}, a large 3D model dataset, to learn the prior knowledge of varying geometries of commonly seen objects and shows good generalization ability on other object datasets. Leveraging this powerful tool, we use the generated meshes as geometric priors for one-shot view planning introduced next. 

\subsection{One-Shot View Planning as Set Covering Optimization} \label{S:our_approach:SCOP}

One-shot view planning can be treated as a conventional set covering optimization problem. Since solving this optimization problem necessitates an explicit 3D representation of the object to be reconstructed, previous works~\citep{pan2022ral1, hu2024icra} rely on depth sensors to acquire initial 3D models of the object. Instead, by incorporating the geometric priors of 3D diffusion models into our planning pipeline, our approach solves the one-shot view planning problem in an RGB camera setup.

To facilitate the efficiency of set covering optimization, sparse surface representations are desired. To this end, we first sample a set of surface points from the mesh produced by the 3D diffusion model and subsequently voxelize them using OctoMap~\citep{hornung2013ar} to get a sparse surface point set $\mathcal{P}_{\mathit{surf}}$, with surface point $p_i \in \mathcal{P}_{\mathit{surf}}$. We denote $v$ as a candidate view within a discrete candidate view space $\mathcal{V} \subset \mathbb{R}^3 \times SO(3)$ and $\mathcal{P}_v$ as the set of surface points observable from this view. Each set $\mathcal{P}_v$ is determined via the ray-casting process implemented in OctoMap. We define an indicator function $I(p,v)$ to represent whether a surface point $p$ is observable from view $v$:

\begin{equation}
\label{equ:indicator}
    I(p,v) = 
    \begin{cases}
        1   & \text{if } p \in \mathcal{P}_v \\
        0   & \text{otherwise}
    \end{cases}
    \, .
\end{equation}

Given $\mathcal{P}_{\mathit{surf}}$ and each $\mathcal{P}_v$, the conventional set covering optimization problem aims to find the minimum set of views required for completely covering the surface points. For instance, consider $\mathcal{P}_{\mathit{surf}} = \left\{p_1,p_2,p_3\right\}$, $\mathcal{P}_{v_1} = \left\{p_1,p_2\right\}$, $P_{v_2} = \left\{p_2,p_3\right\}$, and $\mathcal{P}_{v_3} = \left\{p_1,p_3\right\}$. The union of these three sets equals the entire surface set, i.e., $\bigcup_{v}\mathcal{P}_v = \mathcal{P}_{\mathit{surf}}$. However, we can cover all surface points with only two sets, $\mathcal{P}_{v_1}$ and $\mathcal{P}_{v_2}$.
Vanilla set covering optimization problem requires that each surface point should be covered by at least one view. This definition aligns well with object reconstruction employing depth-sensing modalities~\citep{pan2022ral1, pan2023arxiv, hu2024icra}, as surfaces can be recovered by direct depth fusion when provided with a corresponding point cloud observation. However, for RGB-based object reconstruction using NeRFs, map representation learning is achieved by minimizing the photometric loss when reprojecting hypothetical surface points back to 2D image planes, which requires that a surface point should be observed from different perspectives to recover its true 3D representation. This implies that planned views covering all surface points of the generated mesh once are not sufficient for object reconstruction using NeRFs.

\begin{figure}[!t]
\centering
\includegraphics[width=1.0\columnwidth]{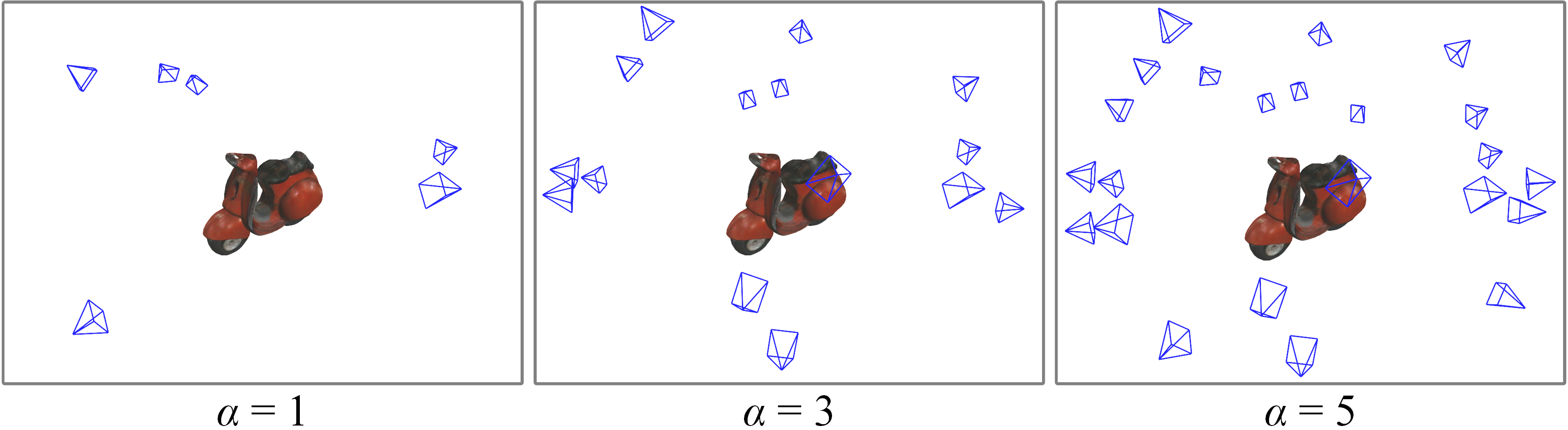}
\caption{
Illustration of the impact of multi-view constraints. $\alpha$ denotes the minimum number of views required to observe each surface point. Larger $\alpha$ values lead to optimization solutions with more views densely covering the surfaces.} 
\label{fig_miniuxm_covering}
\vspace{-0.5cm}
\end{figure}

\begin{figure}[!t]
\centering
\includegraphics[width=0.75\columnwidth]{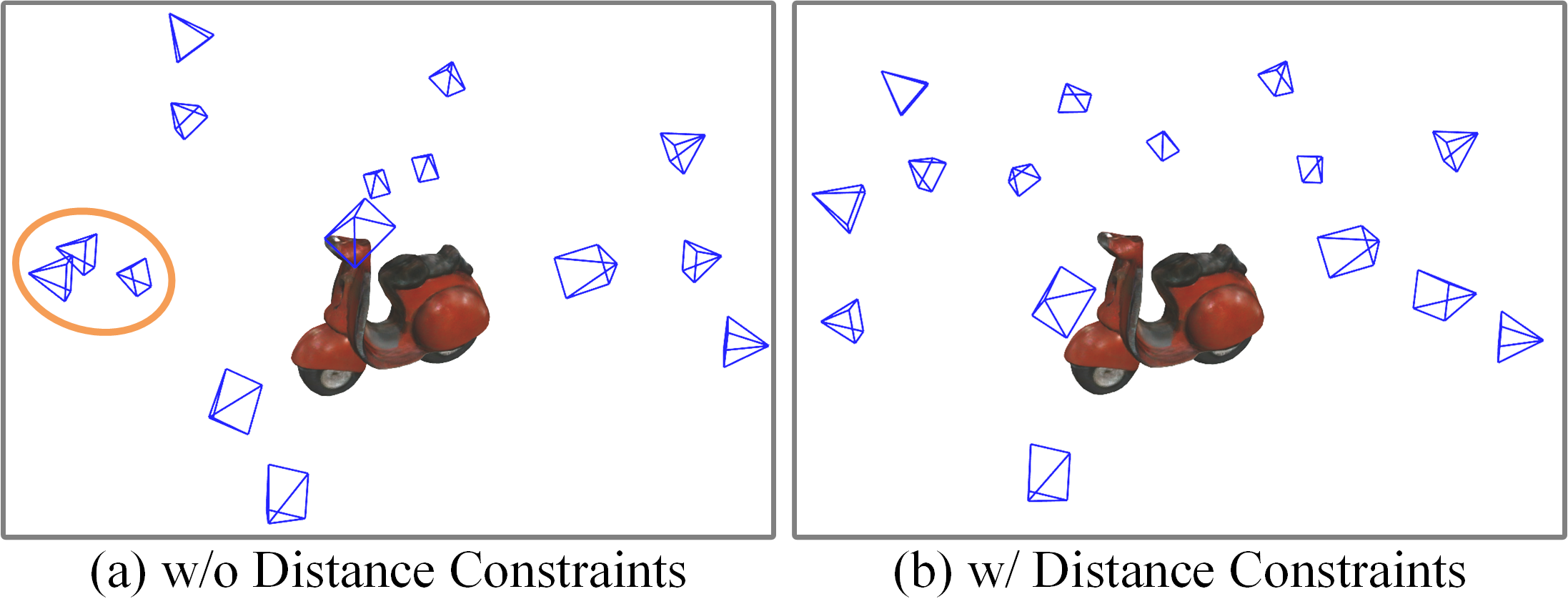}
\caption{
Illustration of the impact of distance constraints: (a) spatially clustered views (the orange circle showcases an example of clustered views); (b) spatially more uniform views. Both view configurations are feasible solutions. By incorporating distance constraints, we express the preference for spatially uniform distribution to avoid redundant information in clustered views.} 
\label{fig_spatial_constraints}
\vspace{-0.5cm}
\end{figure}

To this end, we customize the set covering optimization problem for RGB-based object reconstruction using NeRFs. Rather than requiring each surface point to be observed by at least one view, we propose multi-view constraints to enforce that a given surface point should be covered by a minimum number $\alpha \in \mathbb{N}^{+}$ of views to account for multi-view learning in NeRFs. Larger $\alpha$ values require denser surface coverage in our optimization problem, resulting in solutions with more views required as shown in Fig.~\ref{fig_miniuxm_covering}.
Note that when $\alpha \geq 2$, we exclude points that are visible from fewer than $\alpha$ views. This mechanism ensures the optimization problem has a feasible solution.
However, our multi-view covering setup may contain multiple feasible solutions since most of the surface points can be observed from a large range of view perspectives. Some of them lead to views clustered closely together in Euclidean space. Fig.~\ref{fig_spatial_constraints}(a) illustrates an instance of spatially clustered views for covering the Motorbike object. These spatially clustered views exhibit similarity in the collected images, thus leading to redundant information about the object.

To alleviate this issue, we introduce a parameter \mbox{$\beta \in \mathbb{R}^{\geq 0}$} for additional distance constraints to avoid selecting spatially clustered views. We denote $d_v^{v'}$ as the Euclidean distance between views $v$ and $v'$, while $d_v^{min}$ is the Euclidean distance from view $v$ to its nearest neighboring view. We prevent other views within a specific distance $\beta \, d_v^{min}$ of the view $v$ from being selected again in the solution. A larger $\beta$ leads to more spatially uniform views, while an excessively large value can render the problem infeasible. For our view planning, we try to find the maximum $\beta$ value that still yields an optimization solution. 
Given that different objects exhibit diverse geometries, their respective maximum $\beta$ values also vary. Therefore, we run optimization iteratively to find the maximum $\beta$ for a specific object in an automatic manner.

Taking all these conditions into account, we formulate our set covering optimization problem as a constrained integer linear programming problem defined as follows:

\vspace{-0.2cm}
\begin{equation}
\label{equ:ILP}
\begin{aligned}
\min: \quad & \sum_{v\in \mathcal{V}} x_v \, ,\\
\mathrm{s.t.}: \quad
& (a) \quad x_{v} \in \{0,1\}   && \forall v \in \mathcal{V} \\
& (b) \quad \sum_{v \in \mathcal{V}} I(p, v)\, x_{v} \geq \alpha  && \forall p \in \mathcal{P}_{\mathit{surf}} \\
& (c) \quad x_v + x_{v'} \leq 1  && \forall d_v^{v'} \leq\beta\, d_v^{min},
\end{aligned}
\end{equation}
where the objective function $\sum_{v \in \mathcal{V}} x_v$ is designed to minimize the total number of selected views, while subject to three constraints: (a) $x_{v}$ is a binary variable representing whether a view $v$ is included in the set of selected views or not; (b) each surface point $p \in \mathcal{P}_{\mathit{surf}}$ must be observed by a minimum of $\alpha$ selected views; and (c) if a view $v$ is selected, any neighboring view $v'$, whose distance $d_v^{v'}$ is smaller than $\beta \, d_v^{min}$, must not be selected. 

We employ the Gurobi optimizer, a linear programming solver~\citep{gurobi2021gurobi}, to compute the solution for the problem. We present an instance solution in Fig.~\ref{fig_spatial_constraints}(b) showcasing the optimized minimum set of views required for densely covering the Motorbike object surface with $\alpha=3$ and distance constraints.

\begin{table*}[!t]
\centering
\resizebox{0.87\textwidth}{!}{
\begin{tabular}{cccccc}
\hline
$\alpha$            & Planned Views & \quad\, PSNR $\uparrow$       & \quad\, SSIM $\uparrow$        & \quad Movement Cost (m) $\downarrow$ & \quad Inference Time (s) $\downarrow$ \\ \hline
1 & \, 6.8 ± 1.5       & \, 30.167 ± 0.810 & \, 0.9365 ± 0.0121 & 1.754 ± 0.258    & 140.4 ± 26.9         \\
2 & 12.8 ± 1.7      & \, 31.436 ± 0.622 & \, 0.9530 ± 0.0049 & 2.629 ± 0.224    & 145.9 ± 29.3         \\
3 & 17.8 ± 2.4      & \, 31.853 ± 0.615 & \, 0.9599 ± 0.0038 & 2.998 ± 0.225    & 147.9 ± 31.8         \\
4 & 22.5 ± 3.8      & \, 31.995 ± 0.684 & \, 0.9633 ± 0.0035 & 3.214 ± 0.372    & 148.2 ± 33.1         \\
5 & 28.7 ± 3.8      & \, 32.120 ± 0.786 & \, 0.9663 ± 0.0034 & 3.725 ± 0.312    & 150.0 ± 40.6         \\
6 & 34.1 ± 5.1      & $^\star$32.243 ± 0.779 & $^\star$0.9684 ± 0.0042 & 4.093 ± 0.441    & 147.6 ± 34.1         \\
7 & 38.8 ± 3.8      & $^\dag$32.248 ± 0.807 & $^\dag$0.9694 ± 0.0041 & 4.190 ± 0.247    & 147.3 ± 38.2         \\ \hline
\end{tabular}
}
\caption{Analysis on multi-view constraints. $\alpha$ denotes the minimum number of views required to observe each surface point. Planned views indicate the number of optimized views under different $\alpha$ values. PSNR and SSIM are averaged over $100$ novel views. Each value reports the average mean and standard deviation on $10$ test objects. The star symbol ($\star$) indicates statistically significant results for $\alpha=6$ compared to $\alpha=5$ based on the paired \textit{t}-test with a \textit{p}-value of $0.05$. Conversely, the dagger symbol ($\dag$) indicates non-significant results for $\alpha=7$ compared to $\alpha=6$ based on the paired \textit{t}-test with a \textit{p}-value of $0.05$. Results show that our optimizer plans more views with increasing $\alpha$ values and achieves peak performance at the $\alpha=6$. It is worth mentioning that increasing $\alpha$ from $1$ to $2$ leads to the highest performance gain, indicating that our formulation of set covering benefits NeRF-based reconstruction.
}
\label{tab_performance_ceiling}
\vspace{-0.2cm}
\end{table*}

\subsection{Path Generation and Object Reconstruction}

By planning all required views before data collection, the one-shot view planning paradigm shows a major advantage in reduced movement costs. Given the optimized set of views introduced above, we plan the globally shortest path connecting all views by solving the shortest Hamiltonian path problem on a graph, which is similar to the traveling salesman problem~\cite{osswald2016ral}. The robot's RGB camera follows the global path to acquire RGB measurements at planned views. We follow the point-to-point local path planning method~\cite{pan2023arxiv} to avoid collisions with the object.

After data collection, we use NeRFs to acquire the final 3D representation of the object. Specifically, we adopt \mbox{Instant-NGP~\citep{muller2022tog}} to train our NeRF, due to its efficient training performance and common usage in baseline approaches~\citep{sunderhauf2023icra, lin2022rssworkshop, pan2024icra}.

\section{Experimental Results} \label{S:experimental_results}


\subsection{Experimental Setup}

\begin{figure}[!t]
\centering
\includegraphics[width=0.9\columnwidth]{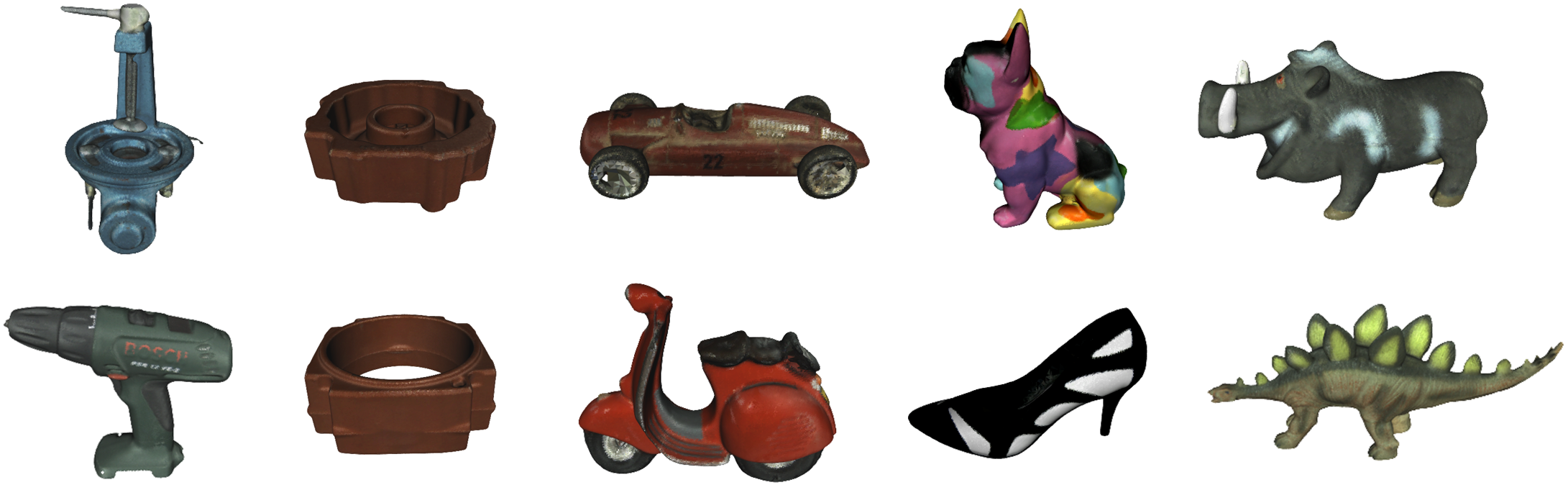}
\caption{
Ten test objects used in our simulation experiments.
} 
\label{fig_object_3D_models}
\vspace{-0.2cm}
\end{figure}

In our simulation experiments, we consider an object-centric hemispherical view space with 144 uniformly distributed view candidates for view planning~\cite{pan2024icra}. We set the view space radius to 0.3\,m.
We test our approach on $10$ geometrically complex 3D object models from the HomebrewedDB dataset~\citep{kaskman2019cvpr}. The test objects are shown in Fig.~\ref{fig_object_3D_models}. We normalize all objects to fit into a bounding sphere with a radius of 0.1\,m. All RGB measurements are at $640$\,px $\times$ $480$\,px resolution. We adopt a grid size of $50 \times 50 \times 50$ in OctoMap for voxelizing the mesh surface points. The set covering optimization for view planning runs on an Intel i7-12700H CPU, while NeRFs training is conducted on an NVIDIA RTX3060 laptop GPU.

To evaluate NeRF reconstruction quality, we report the peak signal-to-noise ratio (PSNR) and the structural similarity index (SSIM)~\cite{mildenhall2020eccv} on 100 uniformly distributed novel views~\cite{pan2024icra}. Additionally, we evaluate reconstruction efficiency by inference time for view planning and accumulated movement cost for data collection in Euclidean distance.

\subsection{Analysis on Multi-View Constraints} \label{S: multiview constraints}

In this section, we explore the influence of multi-view constraints introduced in \mbox{Sec.~\ref{S:our_approach:SCOP}}. We test our methods across varying $\alpha$ values from $1$ to $7$, as detailed in TABLE~\ref{tab_performance_ceiling}. The outcomes reveal that: (1) with increasing $\alpha$ values, our optimizer outputs on average more views for covering the mesh surfaces; (2) both PSNR and SSIM metrics exhibit a consistent improvement with increasing $\alpha$. Specifically, we achieve the highest performance gain by changing $\alpha=1$ to $\alpha=2$, justifying our modification of the set covering optimization problem to account for RGB-based object reconstruction using NeRFs; (3) our method reaches its peak performance at the $\alpha$ value of 6, while increasing~$\alpha$ to a higher value does not yield a statistically significant performance improvement.

\begin{figure}[!t]
\centering
\includegraphics[width=1.0\columnwidth]{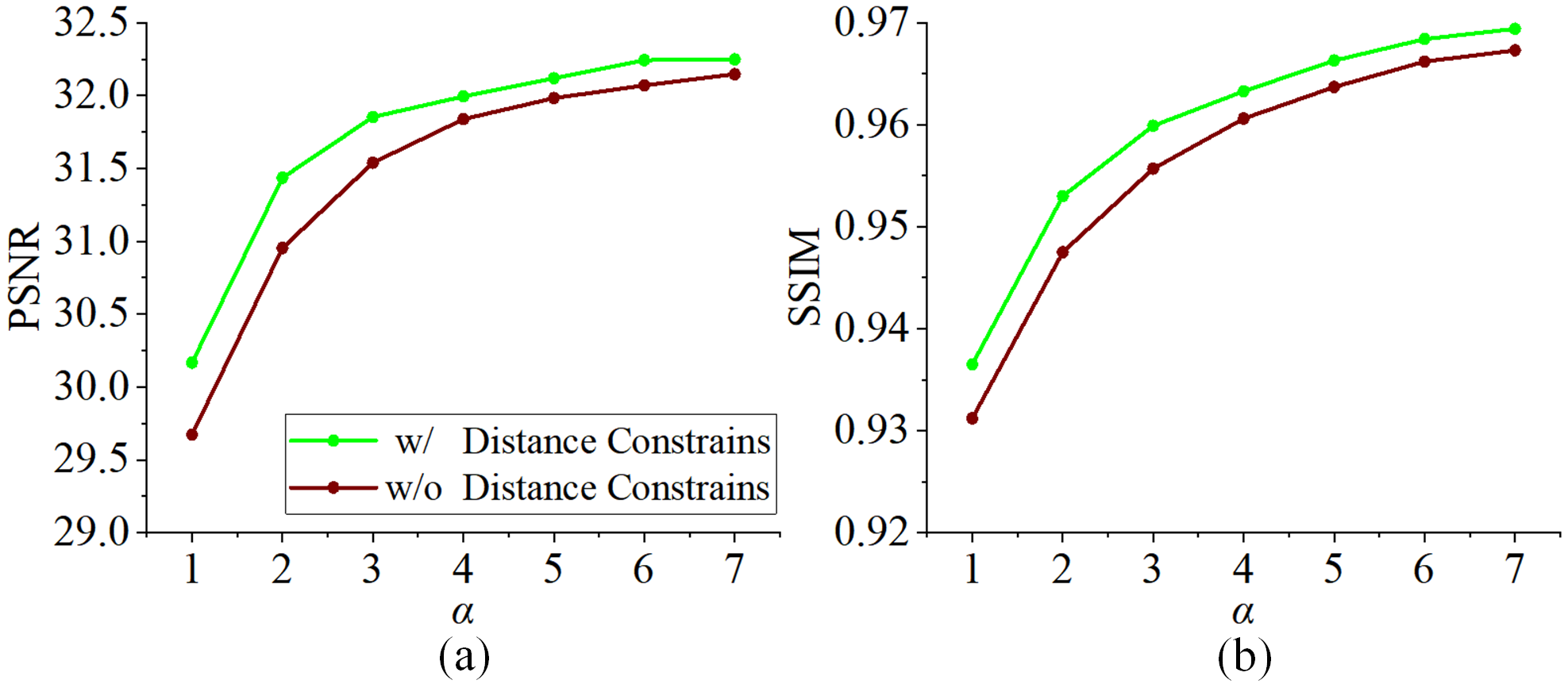}
\caption{
Ablation study on distance constraints. PSNR and SSIM averaged over $100$ novel views. Each value is reported as the averaged mean on $10$ test objects. We observed statistically significant results for our method when compared to the version without distance constraints across all $\alpha$ values, as determined through paired \textit{t}-tests with a \textit{p}-value of $0.05$. This suggests that the set covering optimization with the distance constraints finds better view configurations, leading to superior NeRF training results.
} 
\label{fig_ablation_beta}
\vspace{-0.2cm}
\end{figure}

\subsection{Ablation Study on Distance Constraints}

In this ablation study, we investigate the impact of the distance constraints introduced in Sec.~\ref{S:our_approach:SCOP}. To prevent the optimizer from finding a view configuration that leads to clustered views, we introduce the parameter $\beta$ as the distance constraints into our optimization formulation. We adopt binary search in our implementation to find out the object-specific maximum $\beta$ that still yields a feasible optimization solution. The search step is set to 0.1 for all experiments.

\begin{figure*}[!t]
\centering
\includegraphics[width=1.0\textwidth]{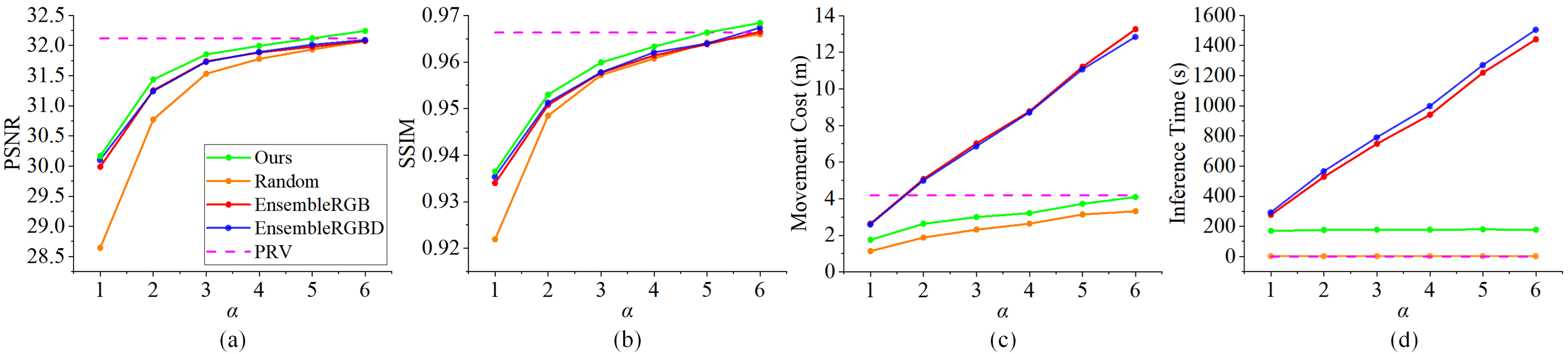}
\caption{
Comparison to baselines on view planning performance under different $\alpha$ values corresponding to the number of optimized views. PSNR and SSIM are averaged over $100$ novel views. Each value reports the mean on $10$ test objects. PRV is not associated with $\alpha$ values and is represented by a dashed line. As can be seen, (1) our method achieves higher PSNR/SSIM values against \textit{Random} and NBV methods, indicating that leveraging geometric priors from diffusion models leads to more informative views; (2) compared to PRV using fixed view configuration, our object-specific view configuration is more suitable for view planning, achieving either a lower movement cost with an on-par performance ($\alpha=5$) or a higher performance with a slightly lower movement cost of 0.09\,m ($\alpha=6$).
} 
\label{fig_baselines_compare}
\end{figure*}

We evaluate the influence of our distance constraints by performing an ablation study over different~$\alpha$ values. Fig.~\ref{fig_ablation_beta} shows the differences between optimization with and without the proposed constraints. In all circumstances, optimization without considering the distance constraints ($\beta=0$) outputs clustered views with redundant information about the object, leading to
inferior NeRF training performance in terms of PSNR and SSIM. This justifies our design choice of introducing the distance constraints to find better view configurations.

\subsection{Evaluation of View Planning for Object Reconstruction} \label{S: view planning}

\textbf{Baselines.} We compare our novel one-shot view planning with the following baselines:
\begin{itemize}
    \item \textit{Random} selects a certain number of views randomly and subsequently plans a global path to connect them.
    \item \textit{EnsembleRGB}~\citep{lin2022rssworkshop} leverages RGB variance of the NeRF ensemble as uncertainty quantification to plan the NBV that maximize the information gain.
    \item \textit{EnsembleRGBD}~\citep{sunderhauf2023icra} extends EnsembleRGB by incorporating a density-aware epistemic uncertainty computed on ray termination probabilities in unobserved object areas.
    \item \textit{PRV}~\citep{pan2024icra} uses a network to predict the required number of views that achieves the peak performance of NeRF training. A fixed hemispherical view configuration is then generated according to the predicted number of views.  
\end{itemize}

For a fair comparison, we use Instant-NGP \citep{muller2022tog} with the same configuration for the training and testing in all experiments. Therefore, the performance differs purely as a consequence of collected RGB images using different planning strategies. As depicted in TABLE~\ref{tab_performance_ceiling}, varying $\alpha$ values result in different numbers of planned views. Therefore, to comprehensively assess the performance of our planner, we evaluate all baselines using an equivalent number of views corresponding to each $\alpha$ value in our approach (excluding PRV, which predicts its own required number of views). 

\textbf{Comparison to Random Selection.} As shown in Fig.~\ref{fig_baselines_compare}, our RGB-based one-shot view planning approach surpasses the one-shot \textit{Random} baseline across all $\alpha$ values in terms of PSNR and SSIM. This is because the heuristic \textit{Random} method does not utilize any available information about the objects, in contrast to our approach. The \textit{Random} method exhibits a slightly lower movement cost. We believe that this occurs since it can produce spatially clustered views, yielding poorer reconstruction quality. These findings confirm that leveraging powerful geometric priors from 3D diffusion models significantly benefits one-shot view planning for RGB-based object reconstruction.

\textbf{Comparison to NBV Methods.} Compared to two NBV baselines, our method achieves higher PSNR and SSIM values across all $\alpha$ values with much less movement costs and inference time, as shown in Fig.~\ref{fig_baselines_compare}. Specifically, our method excels under various resource constraints, e.g., different planned views according to different $\alpha$ values. This implies that using diffusion models for priors leads to more informative views for unknown object reconstruction compared to NBV methods considering the ensemble's variance for uncertainty measurements. We attribute the significant reductions in movement cost and inference time to global path planning and the one-shot non-iterative paradigm, which avoids iterative map updates and uncertainty computation.

\textbf{Comparison to PRV.} Since the PRV method obtains the number of views by predicting the upper limits of NeRF representations, it is not associated with $\alpha$ values and is represented by a dashed line in Fig.~\ref{fig_baselines_compare}. The results indicate that the proposed RGB-based one-shot view planning approach, with an $\alpha = 5$ setting, delivers nearly identical quality metrics in PSNR and SSIM when compared to PRV, yet it benefits from reduced movement cost. Moreover, when $\alpha$ is adjusted to 6, our method surpasses PRV in terms of PSNR and SSIM quality while still maintaining a slightly lower movement cost. This confirms that our object-specific view configuration is superior to fixed view configurations in PRV for handling varying geometries of objects.

In conclusion, our RGB-based one-shot view planning method demonstrates several advantages over the baselines. By integrating powerful geometric priors from 3D diffusion models, our method effectively leverages available object information, resulting in more informative and better distributed views. Moreover, our approach showcases superior adaptability through its object-specific view configuration mechanism. Unlike the fixed view configuration in PRV, our method dynamically adjusts the view configurations for different objects based on their varying geometries. However, we observe a longer inference time of our method compared to the PRV and \textit{Random} methods, primarily due to the constraints imposed by the generation process of the diffusion model (about $60$\,s) and the online optimization process (about $80$\,s). We plan to improve this in the future.

\begin{figure}[!t]
\centering
\includegraphics[width=1.0\columnwidth]{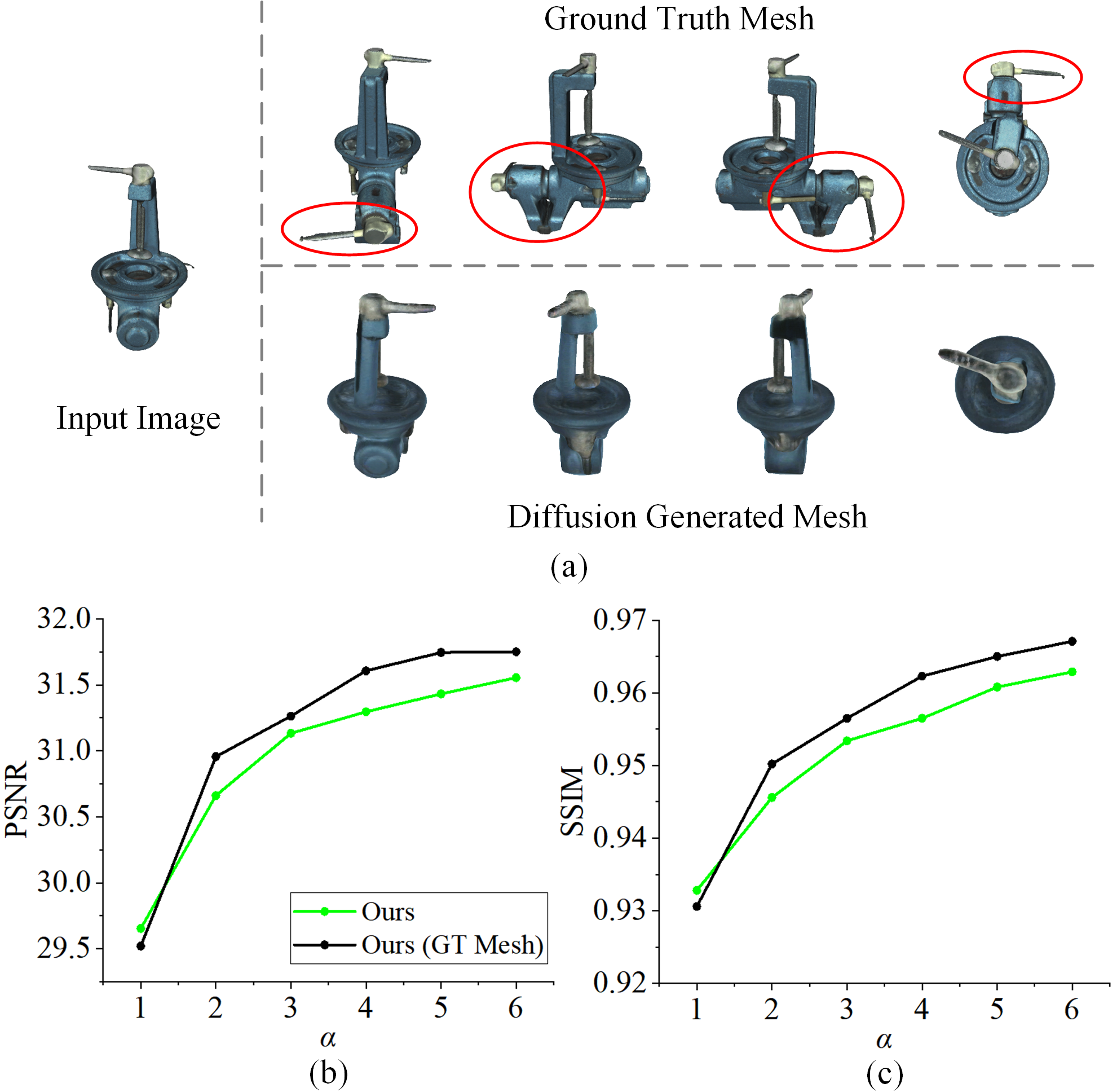}
\caption{
Analysis of a failure case. Top: Input image to the diffusion model and the generated mesh observed from different perspectives. Red circles indicate the missing parts of the generated mesh compared to the ground truth model. Bottom: We compare the reconstruction results using our one-shot view planning based on the ground truth mesh and the generated mesh, showing that wrongly generated geometry leads to reduced performance.
} 
\label{fig_failure_case}
\end{figure}

\subsection{Analysis of Failure Case}

Although our approach successfully performs one-shot view planning from a single RGB image and achieves promising unknown object reconstruction performance, we observe performance inadequacies in a test case. Specifically, the generated mesh from the 3D diffusion model of the Drill object, as depicted in Fig.~\ref{fig_failure_case}(a), demonstrates geometrical discrepancies compared to the ground truth. These disparities might stem from the limited information available due to occlusion in a single input image and the insufficient representation of this type of object in the training dataset. To further validate the impact of this issue on the reconstruction, we conducted experiments by replacing the generated mesh with the ground truth mesh. Fig.~\ref{fig_failure_case}(b-c) reveals that our method using the ground truth mesh achieves higher PSNR and SSIM compared to input with the diffusion-generated mesh. The results indicate that the quality of geometric priors, i.e., the mesh generated from diffusion models, is crucial for our one-shot view planning performance. 

\begin{figure}[!t]
\centering
\includegraphics[width=1.0\columnwidth]{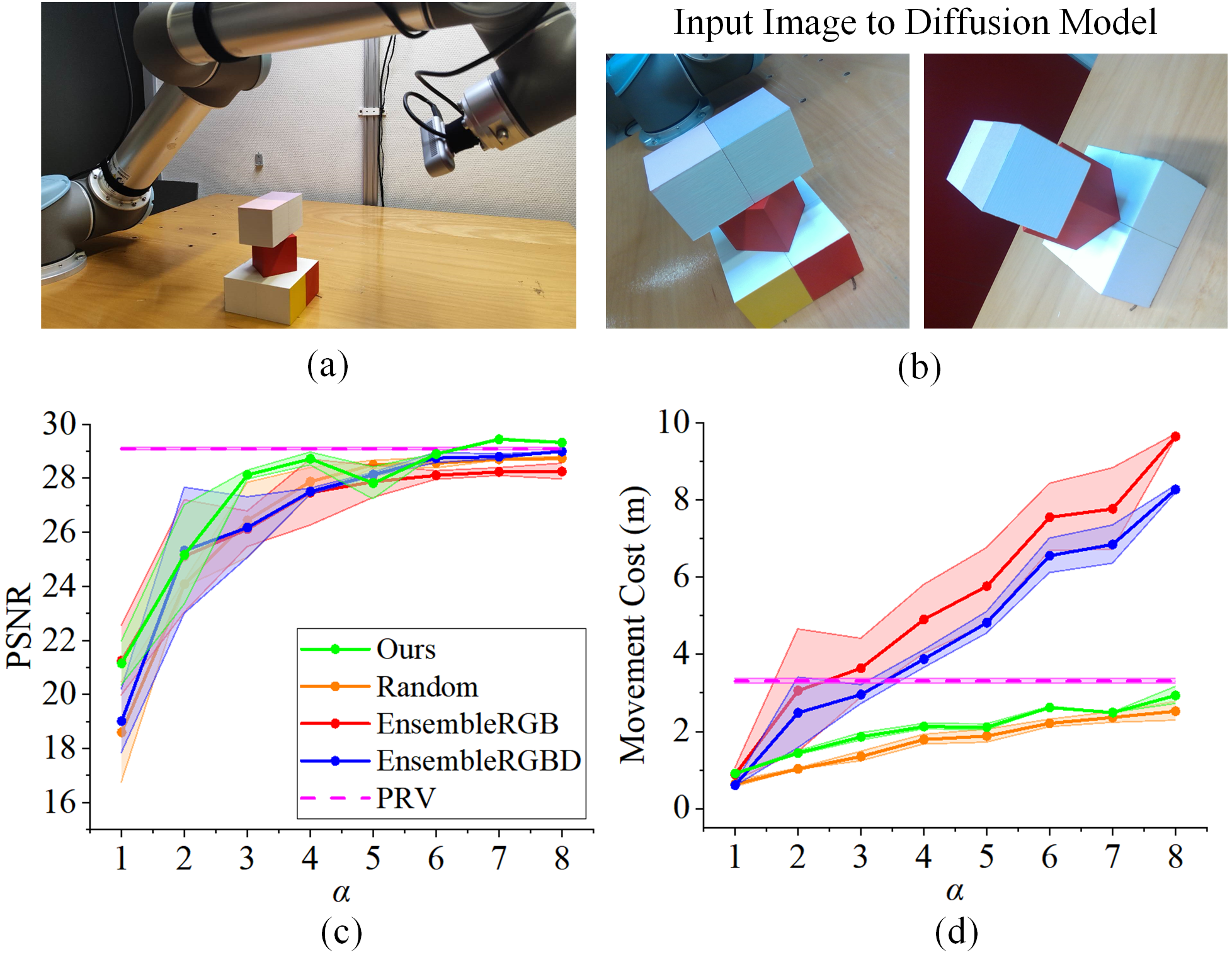}
\caption{
Real-world experiment showing the test object. We run two test trials with different initial views. PSNR is averaged over $100$ novel views. Each value is reported as the averaged mean and with standard deviation (the error bar) on two test trials. By adapting views based on the object geometries, our method achieves a higher performance with lower movement costs.
} 
\label{fig_real_world}
\end{figure}

\subsection{Real-World Experiments}

We deploy our approach in a real world tabletop environment using a UR5 robot arm with an Intel Realsense~D435 camera mounted on its end-effector (only the RGB optical camera is activated). MoveIt~\citep{chitta2016ros} is employed for robotic motion planning. The accompanying video\footnote{\url{https://youtu.be/EKZPHb5-UZk}} illustrates the online reconstruction process where $\alpha=7$.

To validate our findings in Sec.~\ref{S: view planning}, we compare our method against baselines in the real world. It is worth noting that due to imperfect camera poses and noise in real world experiments, the pose optimization functionality implemented in Instant-NGP is enabled during our NeRF training. The experimental environment and comparisons are shown in Fig.~\ref{fig_real_world}. From the results, we confirm that (1) our method generalizes to real world environments; and (2) our method plans object-specific view configurations according to object geometries to achieve higher PSNR with lower movement costs compared to the PRV and NBV methods. Our method achieves peak performance at $\alpha=7$, which is larger than the value of $6$ determined in Sec.~\ref{S: multiview constraints}. This might be caused by the noise in the camera pose and images, making it challenging for view planning tasks. We observe a similar slight performance reduction for the NBV methods.

Nevertheless, when deployed in real-world environments, an estimate of the actual object size is necessary to scale the diffusion-generated models, given that the generated mesh lacks scale information.

\section{Conclusions} \label{S:conclusions}

In this paper, we present a novel one-shot view planning method starting with only a single RGB image of the unknown object to be reconstructed. The proposed method exploits priors from 3D diffusion models as a proxy to the inaccessible ground truth 3D model as the basis for one-shot view planning. We develop a customized variant of the set covering optimization problem tailored for NeRF-based reconstruction, which aims to compute an object-specific view configuration that densely covers the generated mesh from 3D diffusion models. We compute a globally shortest path on this view configuration, corresponding to the minimum travel distance. Our experiments validate that utilizing geometric priors from 3D diffusion models yields more informative views compared to Random and next-best-view methods. When compared to the state-of-the-art RGB-based one-shot baseline, our view planning based on varying object geometries demonstrates better performance compared to a fixed view configuration. The real world experiment suggests the applicability of our method.

\bibliographystyle{IEEEtranSN}
\footnotesize
\bibliography{iros2024}

\end{document}